\documentclass{article}

\usepackage{arxiv}

\usepackage[utf8]{inputenc} 
\usepackage[T1]{fontenc}    
\usepackage{hyperref}       
\usepackage{url}            
\usepackage{booktabs}       
\usepackage{amsfonts}       
\usepackage{nicefrac}       
\usepackage{microtype}      
\usepackage{lipsum}
\usepackage{graphicx}

\title{Machine Learning Methods for Anomaly Detection in Nuclear Power Plant Power Transformers}

\author{
 Iurii Katser \\
  Skolkovo Institute of Science and Technology\\
  Moscow, Russian Federation, 143026 \\
  \texttt{Iurii.Katser@skoltech.ru} \\
  \And
 Dmitriy Raspopov \\
  Cifrum – Nuclear Industry Digitalization, Private Enterprise\\
  Moscow, Russian Federation, 115230 \\
  \texttt{dalraspopov@rosatom.ru} \\
  \And
 Vyacheslav Kozitsin \\
  Skolkovo Institute of Science and Technology\\
  Moscow, Russian Federation, 143026 \\
  \texttt{Vyacheslav.Kozitsin@skoltech.ru} \\
  \And
 Maxim Mezhov \\
  Digital Technologies and Platforms, LLC\\
  Moscow, Russian Federation, 115054 \\
  \texttt{msmezhov@ya.ru} \\
}

\begin{document}
\maketitle
\begin{abstract}
Power transformers are an important component of a nuclear power plant (NPP). Currently, the NPP operates a lot of power transformers with extended service life, which exceeds the designated 25 years. Due to the extension of the service life, the task of monitoring the technical condition of power transformers becomes urgent.
An important method for monitoring power transformers is Chromatographic Analysis of Dissolved Gas. It is based on the principle of controlling the concentration of gases dissolved in transformer oil. The appearance of almost any type of defect in equipment is accompanied by the formation of gases that dissolve in oil, and specific types of defects generate their gases in different quantities.
At present, at NPPs, the monitoring systems for transformer equipment use predefined control limits for the concentration of dissolved gases in the oil.
This study describes the stages of developing an algorithm to detect defects and faults in transformers automatically using machine learning and data analysis methods. Among machine learning models, we trained Logistic Regression, Decision Trees, Random Forest, Gradient Boosting, Neural Networks. The best of them were then combined into an ensemble (\texttt{StackingClassifier}) showing F1-score of 0.974 on a test sample. To develop mathematical models, we used data on the state of transformers, containing time series with values of gas concentrations (H\textsubscript{2}, CO, C\textsubscript{2}H\textsubscript{4}, C\textsubscript{2}H\textsubscript{2}). The datasets were labeled and contained four operating modes: normal mode, partial discharge, low energy discharge, low-temperature overheating.
\end{abstract}

\keywords{Power transformer \and Nuclear power plant \and Chromatographic Analysis of Dissolved Gas \and Machine learning \and Technical diagnostics \and Anomaly detection \and Diagnosis.}

\section{Introduction}
\label{sec:headings}
Modern nuclear power plants (NPPs) generate a large amount of data. Mining or data-driven methods provide effective using of the generated information for faults and failures detection, deter- mining the remaining useful life of the equipment, and solving other urgent diagnostic tasks. The International Atomic Energy Agency (IAEA) states that the high quality of solving these tasks and effective implementation of modern techniques will allow nuclear industry to switch to condition-based maintenance (CBM) \cite{ref_article1} which has been successfully applied for different applications. According to Bond et al. and their analysis \cite{ref_article8}, applying of the CBM in the nuclear industry will be saving for the United states over 1 billion per year. Moreover, it is strongly important to keep in safe NPPs defending against cyber-attacks and intrusions, equipment faults and failures, and other anomalies to prevent the terrible and devastating consequences. Moreover the downtime of the plant leads to significant financial losses. At the same time, safety requirements need to be complied with cost-effectively without expensive additional diagnostics and safety systems and without adding a large number of new measuring channels. Finally, faults in equipment and processes are stated as one of the challenges in NNPs by IAEA.

Power transformers (PTs) are an important component of a NPP. They convert alternating voltage and are instrumental in power supply of both external NPP energy consumers and NPPs themselves. Currently, many PTs have exceeded planned service life that had been extended over the designated 25 years. Due to the extension, monitoring the PT technical condition becomes an urgent matter.

An important method for monitoring and diagnosing PTs is Chromatographic Analysis of Dissolved Gas (CADG). It is based on the principle of forced extraction and analysis of dissolved gases from PT oil. Almost all types of equipment defects are accompanied by formation of gases that dissolve in oil; certain types of defects generate certain gases in different quantities. At present, NPP control and diagnostic systems for PT equipment use predefined control limits for concentration of dissolved gases in oil. The main disadvantages of this approach are the lack of automatic control and insufficient quality of diagnostics, especially for PTs with extended service life. To combat these shortcomings in diagnostic systems for the analysis of data obtained using CADG, machine learning (ML) methods including neural networks, as well as pattern recognition algorithms, time series forecasting, cluster analysis, control charts, graph analysis, etc can be used. The variety of methods used for NPP and other complex technical systems and their components diagnostics is represented in the IAEA report \cite{ref_article1}, books \cite{ref_article9,ref_article10,ref_article11} and recent reviews \cite{katser2019npp,ref_article13,ref_article14,ref_article15,ref_article16,ref_article17,ref_article18,ref_article19}.
Mentioned ML methods showed significant results in solving various tasks in many industrial applications including anomaly detection task \cite{ref_article12}.

Our study describes the stages of developing an algorithm for detecting and classifying anomalies (defects and faults) in PTs automatically using ML and data analysis methods.

\section{Problem statement}
\label{sec:headings}
\begin{figure}
    \centering
    \includegraphics[scale=0.35]{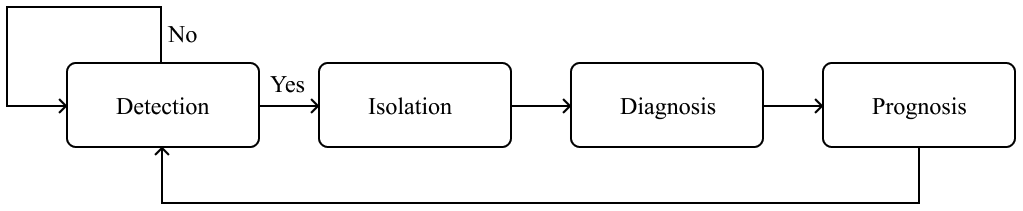}
    \caption{Process monitoring circuit.} \label{fig0}
\end{figure}

The fault detection problem is a particular case of the anomaly detection problem in the data that refers to the monitoring of industrial systems and components which are PTs.
Moreover, fault detection is one of the four diagnostics tasks \cite{ref_article1}, which are illustrated as a sequence of
diagnosis process in Fig.~\ref{fig0}. Fault detection is a critical link in the diagnostic circuit since only in case of a deviation detection in the equipment operation, the processes of solving the remaining diagnostic problems (identification or isolation, diagnosis, prognosis or process recovery) are started.

Depending on existing data and labels, business problems, and desired output, a fault detection problem may transform into various mathematical problems, such as classification, regression, changepoint detection, etc. 
Almost always in technical systems diagnostics and anomaly detection problem in particular there is a lack of markup or labels for the anomalous state. 
We often have only healthy state to train reconstruction or prediction model of the normal state and look for deviations of the model’s outputs on the new data. 
The details regarding these approaches are presented in \cite{geiger2020tadgan}. Sometimes we do not have even healthy state and pure unsupervised techniques are used \cite{shvetsov2020unsupervised,truong2020selective,katser2021unsupervised}.
In our case, the fault detection problem transforms into a classification problem, since the data is related to one of four labeled classes (including one normal and three anomalous), so the model's output needs to be a class number. Since the algorithm solves multiclass classification problem indicating specific state or operation mode, it also solves the diagnosis problem. Moreover, we do not train two different models for fault detection and diagnosis problems solving but we use a single one multiclass classifier for both.

\section{Data}
\label{sec:headings}
One of the challenges for the research of ML application to manufacturing, energy-producing and construction industries is a lack of publicly available non-proprietary real-world data. There are just a few public benchmarks and datasets in technical anomaly detection field:
\begin{itemize}
    \item Tennessee Eastman Process (TEP) benchmark \cite{downs1993plant};
    \item Sugar Refinary Benchmark \cite{Bartys2006};
    \item Urban waste water treatment plant \cite{poch1993faults};
    \item Delft pump \cite{ypma1999pump};
    \item Satellite anomaly benchmark \cite{Satellite};
    \item Wireless Sensor Networks benchmark \cite{suthaharan2010labelled};
    \item Machinery Fault Database \cite{mafaulda};
    \item Skoltech anomaly benchmark (SKAB) \cite{katser2020skoltech}.
\end{itemize}

But none of the mentioned datasets relate to nuclear field, and just a few of them represent real-world data. 

This paper firstly introduces a real-world data with PT anomalies which is publicly available at \url{https://www.kaggle.com/competitions/transformer/data}.
The data represents PT state and contains 3000 separate datasets. Each dataset contains time series with gas concentrations values (H\textsubscript{2}, CO, C\textsubscript{2}H\textsubscript{4}, C\textsubscript{2}H\textsubscript{2}). The datasets are labeled and contain four PT operating modes (classes):
\begin{enumerate}
    \item Normal mode;
    \item Partial discharge: local dielectric breakdown in gas-filled cavities;
    \item Low energy discharge: sparking or arc discharges in poor contact connections of structural elements with different or floating potential; discharges between PT core structural elements, high voltage winding taps and the tank, high voltage winding and grounding; discharges in oil during contact switching;
    \item Low-temperature overheating: oil flow disruption in windings cooling channels, magnetic system causing low efficiency of the cooling system for temperatures $<$ 300 °C.
\end{enumerate}

\begin{figure}
    \centering
    \includegraphics[width=\textwidth]{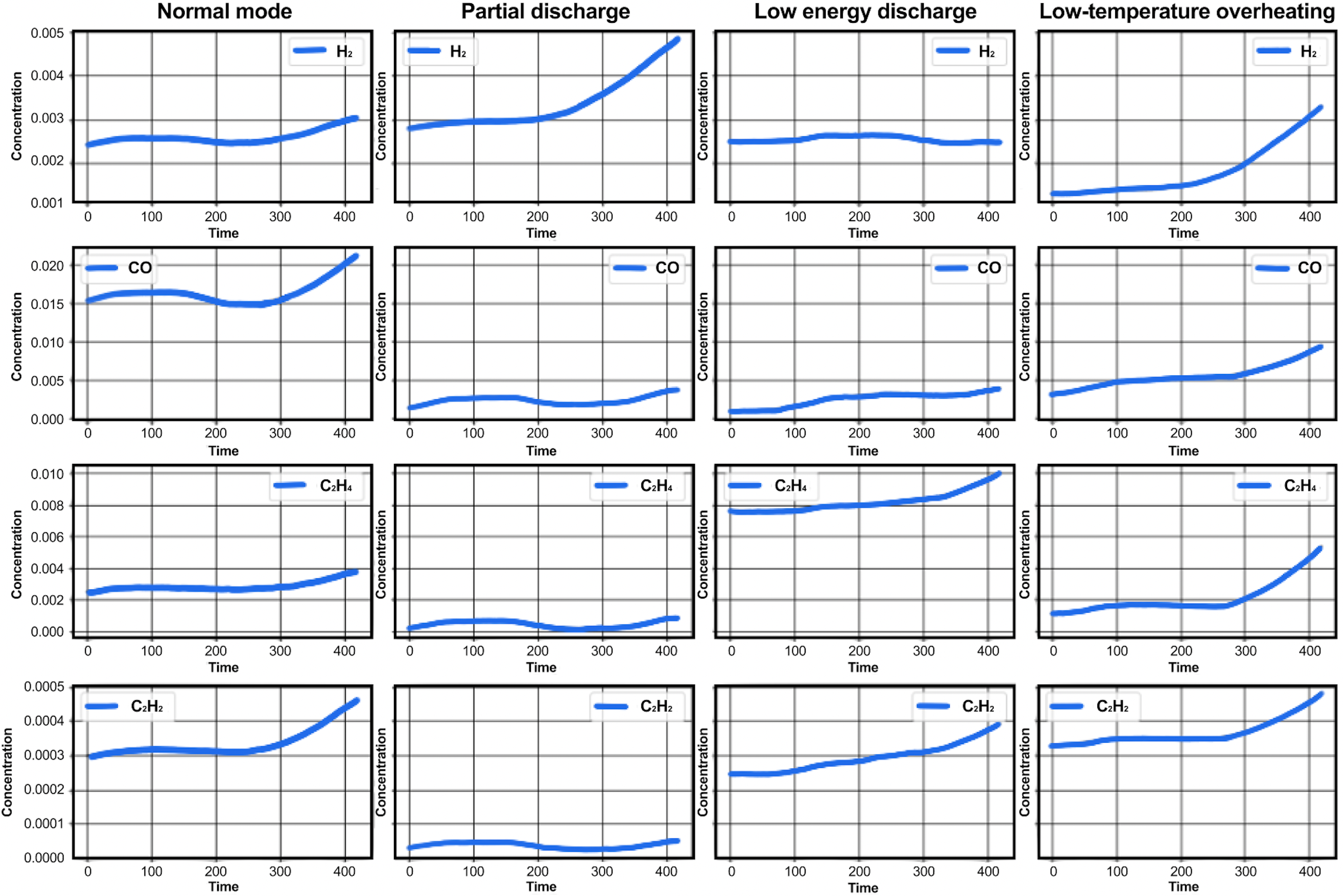}
    \caption{Examples of datasets (columns show various operating modes; rows show various gases).} \label{fig1}
\end{figure}

A single label meaning single operation mode is assigned to each dataset which may be represented in a matrix form of a 4 (features) x 420 (points) size. The raw features represent various gases, while each point is a value of gas concentration measured over twelve hours. Examples of datasets for each class are presented in Fig.~\ref{fig1}).

\begin{figure}
    \centering
    \includegraphics[width=\textwidth]{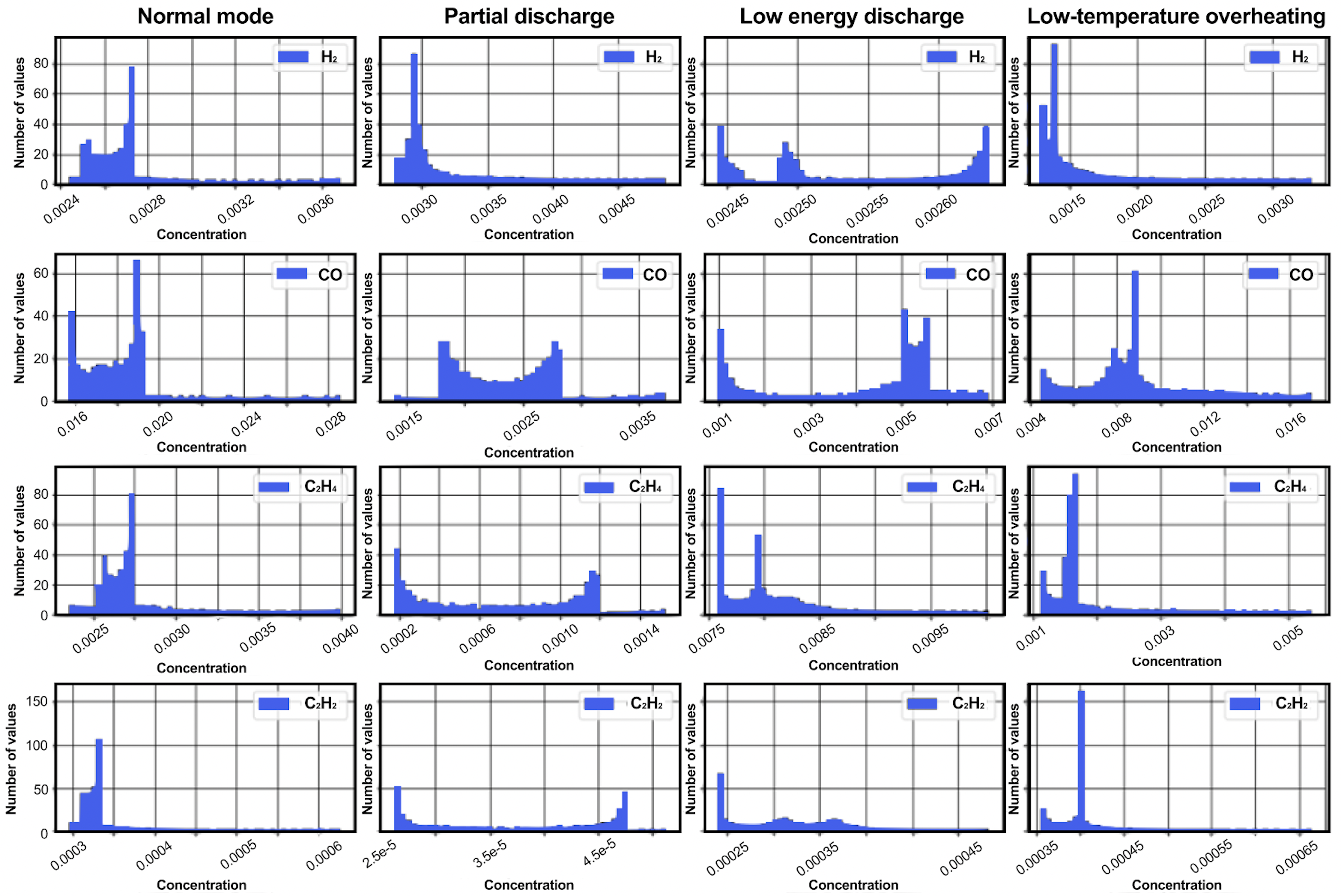}
    \caption{Distributions of datasets (columns show various operating modes; rows show various gases).} \label{fig11}
\end{figure}

Since we have labels for the data with a relatively lot of incidents, supervised techniques are much more preferable due to their high or state-of-the-art results in plenty tasks. Moreover, when looking at the distributions of the same signals in different modes (Fig.~\ref{fig11}), it becomes clear that various modes can be divided according to the values of the statistical properties (mean, std, etc.). This leads to an idea of extracting such features and feeding them to the models. The only problem remains how to preprocess data of time-series type to use classic supervised techniques appropriately and effectively.

\begin{figure}
    \centering
    \includegraphics[scale=0.55]{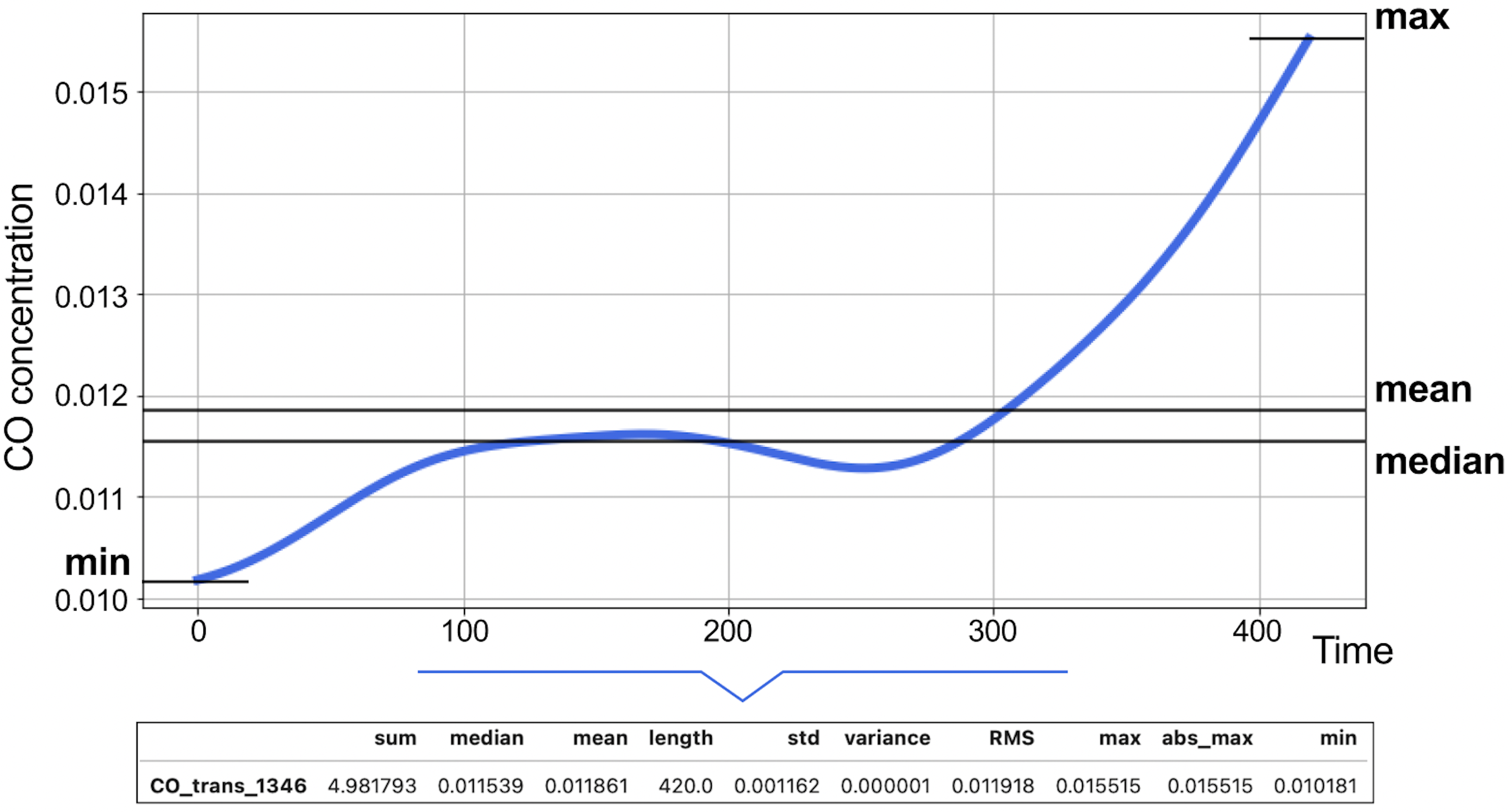}
    \caption{The scheme of time series feature extraction procedure.} \label{fig12}
\end{figure}

To be properly used as the input into ML models, the data needs to be preprocessed and transformed into a single vector, not a matrix. The details about the importance of preprocessing in general and a review of preprocessing stages and methods for NPPs are presented in \cite{ref_article3}. In our case, feature extraction procedure over a time series from \texttt{tsfresh} library \cite{christ2018time} are used. The scheme of these transformations is presented in Fig.~\ref{fig12}.

This procedure contains aggregations that transform each vector to a single value. After several aggregations, several new features representing statistical properties (mean, median, max, min, std, etc.) are acquired. After the full feature extraction procedure, informative features are created and at the next step they are selected by dropping strongly correlated features (Pearson correlation coefficient $>$ 0.9). The result is a 16 (features) x 900 (points) dataset. After that, the dataset is divided into train and test (20\%) sets, and ML models are trained and tested using the preprocessed data.

\section{Methods and Results}

\subsection{Evaluation metrics}
To compare various algorithms in presence of unbalanced classes in data we have chosen Precision, Recall and F1 metrics.
\begin{equation}
Precision =\frac{T P}{T P+F P} \label{eq1}
\end{equation}

\begin{equation}
Recall=\frac{T P}{T P+F N} \label{eq2}
\end{equation}

\begin{equation}
F 1=2 \cdot \frac{ Precision  \cdot Recall }{ Precision + Recall } \label{eq3}
\end{equation}
where \textit{TP} is the total number of \textit{tp} samples; \textit{FP} is the total number of \textit{fp} samples; \textit{FN} is the total number of \textit{fn} samples; \textit{TN} is the total number of \textit{tn} samples.

The intuition behind determining the classification result for each sample by using a confusion matrix is shown in Table~\ref{tab1}.

\begin{table}
\centering
\caption{Confusion matrix.}\label{tab1}
\begin{tabular}{|l|l|l|}
\hline
Actual $\backslash$ Predicted & Predicted Anomaly & Predicted no anomaly \\
\hline
Actual anomaly & true positive (\textit{tp}) & false negative (\textit{fn}) \\
\hline
Actual no anomaly  & false positive (\textit{fp}) & true negative (\textit{tn}) \\
\hline
\end{tabular}
\end{table}

\subsection{Train stage \#1: classic ML methods}
We selected and used during the experiments ML methods of the most common classes, including linear method (Logistic Regression), tree-based method (Decision Trees), ensemble-based methods (Random Forest, Gradient Boosting) and neural network (multilayer perceptron).

A comparison of the quality of the algorithms is shown in Table~\ref{tab12}. The F1 metric on both test set and five folds cross-validation was used to select the best model. For cross-validation, we look at the mean and standard deviation of the metric calculated over the five folds which represent bias and variance respectively. All the ensemble models and neural network show quite similar results. Even though \texttt{RandomForestClassifier} slightly loses to other ensemble models on the cross-validation, the standard deviation of the metric is the smallest among all models.

\begin{table}
\centering
\caption{Metrics of classifiers on test set and 5-folds cross-validation.}\label{tab12}
\begin{tabular}{|l|l|l|l|}
\hline
ML algorithm & F1-score on test set & \multicolumn{2}{c|}{F1-score on 5-folds} \\
 &  & mean & std \\
\hline
\texttt{LogisticRegression} & 0.95 & 0.94 & 0.009 \\
\hline
\texttt{DecisionTreeClassifier} & 0.94 & 0.92 & 0.008 \\
\hline
\texttt{RandomForestClassifier} & 0.96 & 0.94 & 0.006 \\
\hline
\texttt{LGBMClassifier} & 0.96 & 0.95 & 0.011 \\
\hline
\texttt{XGBClassifier} & 0.96 & 0.95 & 0.009 \\
\hline
\texttt{MLPClassifier} & 0.96 & 0.96 & 0.011 \\
\hline
\end{tabular}
\end{table}

\subsection{Train stage \#2: ensemble of selected models (\texttt{StackingClassifier})}

Based on the previous results, we have selected following models to combine them in the single algorithm called \texttt{StackingClassifier}:
\begin{itemize}
    \item \texttt{LGBMClassifier}, an implementation of the gradient boosting model in \texttt{LightGBM} framework \cite{ref_article4};
    \item \texttt{RandomForestClassifier}, an implementation of the random forest model in \texttt{scikit-learn} framework \cite{ref_article5};
    \item \texttt{MLPClassifier}, a multilayer perceptron implemented with \texttt{tensorflow} framework \cite{ref_article6}. In our case, a simple two-layer perceptron was used.
\end{itemize}

A technique called stacking was used to build an ensemble of models. In this technique, heterogeneous individual models can be combined. There is a meta-model, which takes the basic models as input, and meta-model's output is the final prediction of the whole algorithm. It is often improves quality and robustness of the solution \cite{katser2021unsupervised}. In our case, selected models were combined into an ensemble using a second-level gradient boosting model implemented in the \texttt{XGBoost} library \cite{ref_article7} that uses first-level (basic) models' outputs as the features for its input. The training of the \texttt{StackingClassifier} (train stage \#2) is following:
\begin{enumerate}
    \item The train set is divided into K folds;
    \item For an object from the train set, related to the K-th fold, a prediction is made by weak algorithms that have been trained on K-1 folds. This process is iterative and happens for each fold.
    \item A set of basic models' predictions is created for each object in the train set;
    \item The meta-model is trained on the predictions generated by first-level algorithms.
\end{enumerate}

\begin{figure}
\includegraphics[width=\textwidth]{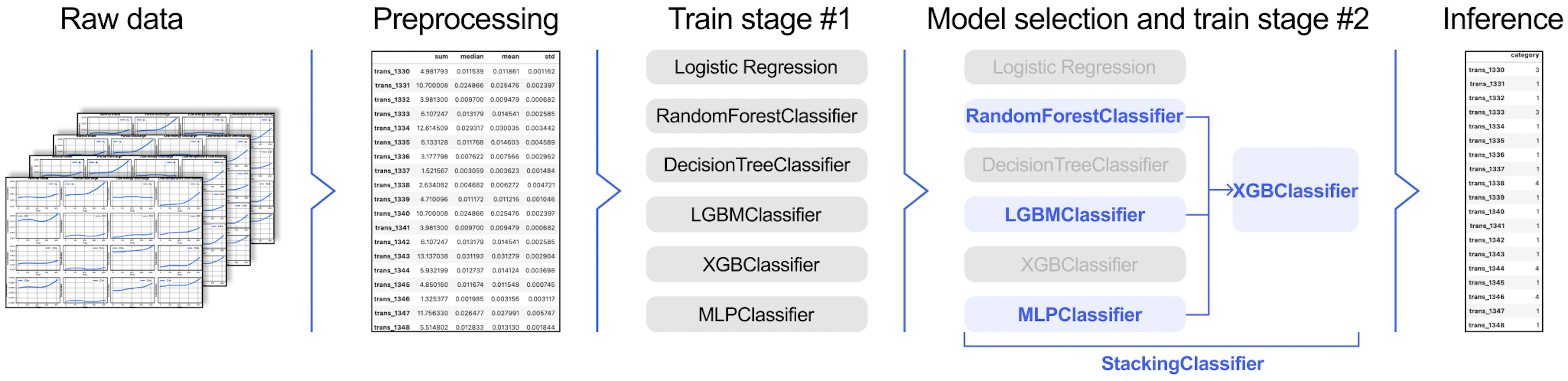}
\caption{The pipeline of the diagnostic approach development.} \label{fig2}
\end{figure}

The final pipeline, as well as the ensembling scheme and ML models used in the final solution are shown in Fig.~\ref{fig2}).

In Table~\ref{tab2}, the detailed results of the classification problem solving including multi-class (multi-label) classification metrics are presented. The details regarding the multi-class classification metrics are given in \cite{grandini2020metrics}.


\begin{table}
\centering
\caption{Detailed results of the \texttt{StackingClassifier} on test set.}\label{tab2}
\begin{tabular}{|l|l|l|l|}
\hline
Class & Precision & Recall & F1 \\
\hline
Normal mode & 0.98 & 0.99 & 0.99 \\
\hline
Partial discharge & 0.91 & 1.0 & 0.95 \\
\hline
Low energy discharge & 0.78 & 0.78 & 0.78 \\
\hline
Low-temperature overheating & 1.0 & 0.89 & 0.94 \\
\hline
Macro average & 0.92 & 0.91 & 0.91 \\
\hline
Weighted average & 0.97 & 0.97 & 0.97 \\
\hline
\end{tabular}
\end{table}

The misclassifications in the form of a confusion matrix for the \texttt{StackingClassifier} model on the test set are shown in Table~\ref{tab23}. Using the \texttt{StackingClassifier} model, a prediction was made on the test set, and the following results were obtained: 5 errors for the Normal mode class, 1 error for the Partial discharge class, 5 errors for the Low-energy discharge class, 7 errors for the Low-temperature overheating class. These results indicate that Low-energy discharge class is the hardest to predict correctly, while both Low-energy discharge class and Low-temperature overheating class misclassifications are falsely assigned to the normal mode class in most cases. Finally, Partial discharge and normal mode classes are the easiest to predict.

\begin{table}
\centering
\caption{Confusion matrix for \texttt{StackingClassifier} on test set.}\label{tab23}
\begin{tabular}{|l|l|l|l|l|}
\hline
Actual / Predicted class & Normal mode & Partial discharge & Low energy discharge & Low-temp. overheating \\
\hline
Normal mode & 342 & 0 & 3 & 2 \\
\hline
Partial discharge & 0 & 15 & 0 & 1 \\
\hline
Low energy discharge & 5 & 0 & 12 & 0 \\
\hline
Low-temp. overheating & 6 & 1 & 0 & 33 \\
\hline
\end{tabular}
\end{table}

\section{Conclusion}

We have described the stages of developing an algorithm for anomaly detection and diagnosis in power transformers using machine learning methods (Fig.~\ref{fig2}). At the initial stage, informative features were created and selected. After that, machine learning models were trained on the preprocessed data. The preprocessing part, including train-test splitting have been described. As a core part of the algorithm, we have trained Logistic Regression, Decision Trees, Random Forest, Gradient Boosting and Neural Network (perceptron). The final part of the algorithm development is the selecting and ensembling the best ML models. Developed algorithm \texttt{StackingClassifier} shows an F1 of 0.974 on a test set. Since the algorithm's output is a class of PT's state, it allows to solve two of the technical diagnostics tasks: anomaly (fault) detection and diagnosis. 

The research also presents the real-world data from PTs that represents the state of transformers, containing time series with values of gas concentrations (H\textsubscript{2}, CO, C\textsubscript{2}H\textsubscript{4}, C\textsubscript{2}H\textsubscript{2}) and one of four modes (normal mode, partial discharge, low energy discharge,
low-temperature overheating) as labels. 

The results clearly show that ML methods are applicable in technical diagnostics and PT monitoring using Chromatographic Analysis of Dissolved Gas. ML methods can also improve the current level of diagnostics. The developed algorithms might improve fault detection and diagnosis process, which leads to an increasing efficiency and potentially allows avoiding major accidents that often result in a lot of fatalities or significant financial losses.

\bibliographystyle{unsrt}  
\bibliography{references}  

\begin{thebibliography}{10}

\bibitem{ref_article1}
{\em Advanced Surveillance, Diagnostic and Prognostic Techniques in Monitoring
  Structures, Systems and Components in Nuclear Power Plants}.
\newblock Number NP-T-3.14 in Nuclear Energy Series. INTERNATIONAL ATOMIC
  ENERGY AGENCY, Vienna, 2013.

\bibitem{ref_article8}
Leonard~J Bond, Pradeep Ramuhalli, Magdy~S Tawfik, and Nancy~J Lybeck.
\newblock Prognostics and life beyond 60 years for nuclear power plants.
\newblock In {\em 2011 IEEE Conference on Prognostics and Health Management},
  pages 1--7. IEEE, 2011.

\bibitem{ref_article9}
Leo~H Chiang, Evan~L Russell, and Richard~D Braatz.
\newblock {\em Fault detection and diagnosis in industrial systems}.
\newblock Springer Science \& Business Media, 2000.

\bibitem{ref_article10}
Douglas~C Montgomery.
\newblock {\em Introduction to statistical quality control}.
\newblock John Wiley \& Sons, 2007.

\bibitem{ref_article11}
Chris Aldrich and Lidia Auret.
\newblock {\em Unsupervised process monitoring and fault diagnosis with machine
  learning methods}, volume~16.
\newblock Springer, 2013.

\bibitem{katser2019npp}
I~Katser, V~Kozitsin, and I~Maksimov.
\newblock Npp equipment fault detection methods.
\newblock {\em Izvestiya vuzov. Yadernaya Energetika}, 4:5--27, 2019.

\bibitem{ref_article13}
Venkat Venkatasubramanian, Raghunathan Rengaswamy, Kewen Yin, and Surya~N
  Kavuri.
\newblock A review of process fault detection and diagnosis: Part i:
  Quantitative model-based methods.
\newblock {\em Computers \& chemical engineering}, 27(3):293--311, 2003.

\bibitem{ref_article14}
Venkat Venkatasubramanian, Raghunathan Rengaswamy, and Surya~N Kavuri.
\newblock A review of process fault detection and diagnosis: Part ii:
  Qualitative models and search strategies.
\newblock {\em Computers \& chemical engineering}, 27(3):313--326, 2003.

\bibitem{ref_article15}
Venkat Venkatasubramanian, Raghunathan Rengaswamy, Surya~N Kavuri, and Kewen
  Yin.
\newblock A review of process fault detection and diagnosis: Part iii: Process
  history based methods.
\newblock {\em Computers \& chemical engineering}, 27(3):327--346, 2003.

\bibitem{ref_article16}
S~Joe Qin.
\newblock Data-driven fault detection and diagnosis for complex industrial
  processes.
\newblock {\em IFAC Proceedings Volumes}, 42(8):1115--1125, 2009.

\bibitem{ref_article17}
Jianping Ma and Jin Jiang.
\newblock Applications of fault detection and diagnosis methods in nuclear
  power plants: A review.
\newblock {\em Progress in nuclear energy}, 53(3):255--266, 2011.

\bibitem{ref_article18}
Dawn An, Joo~Ho Choi, and Nam~Ho Kim.
\newblock Options for prognostics methods: A review of data-driven and
  physics-based prognostics.
\newblock In {\em 54th AIAA/ASME/ASCE/AHS/ASC Structures, Structural Dynamics,
  and Materials Conference}, page 1940, 2013.

\bibitem{ref_article19}
Xuewu Dai and Zhiwei Gao.
\newblock From model, signal to knowledge: A data-driven perspective of fault
  detection and diagnosis.
\newblock {\em IEEE Transactions on Industrial Informatics}, 9(4):2226--2238,
  2013.

\bibitem{ref_article12}
Varun Chandola, Arindam Banerjee, and Vipin Kumar.
\newblock Anomaly detection: A survey.
\newblock {\em ACM computing surveys (CSUR)}, 41(3):1--58, 2009.

\bibitem{geiger2020tadgan}
Alexander Geiger, Dongyu Liu, Sarah Alnegheimish, Alfredo Cuesta-Infante, and
  Kalyan Veeramachaneni.
\newblock Tadgan: Time series anomaly detection using generative adversarial
  networks.
\newblock In {\em 2020 IEEE International Conference on Big Data (Big Data)},
  pages 33--43. IEEE, 2020.

\bibitem{shvetsov2020unsupervised}
Nikolay Shvetsov, Nazar Buzun, and Dmitry~V Dylov.
\newblock Unsupervised non-parametric change point detection in quasi-periodic
  signals.
\newblock {\em arXiv preprint arXiv:2002.02717}, 2020.

\bibitem{truong2020selective}
Charles Truong, Laurent Oudre, and Nicolas Vayatis.
\newblock Selective review of offline change point detection methods.
\newblock {\em Signal Processing}, 167:107299, 2020.

\bibitem{katser2021unsupervised}
Iurii Katser, Viacheslav Kozitsin, Victor Lobachev, and Ivan Maksimov.
\newblock Unsupervised offline changepoint detection ensembles.
\newblock {\em Applied Sciences}, 11(9):4280, 2021.

\bibitem{downs1993plant}
James~J Downs and Ernest~F Vogel.
\newblock A plant-wide industrial process control problem.
\newblock {\em Computers \& chemical engineering}, 17(3):245--255, 1993.

\bibitem{Bartys2006}
Micha{\l} Barty{\'s}, Ron Patton, Micha{\l} Syfert, Salvador de~las Heras, and
  Joseba Quevedo.
\newblock Introduction to the damadics actuator fdi benchmark study.
\newblock {\em Control engineering practice}, 14(6):577--596, 2006.

\bibitem{poch1993faults}
M~Poch, J~Bejar, and U~Cortes.
\newblock Faults in an urban waste water treatment plant (dataset).
\newblock {\em Donated to the UCI Machine Learning Database Repository}, 1993.

\bibitem{ypma1999pump}
A~Ypma, R~Ligteringen, RPW Duin, and EEE Frietman.
\newblock Pump vibration datasets.
\newblock {\em Pattern recognition group, Delft University of Technology},
  1999.

\bibitem{Satellite}
Solar-Terrestrial Physics~Division of~the National Geophysical Data~Center.
\newblock Spacecraft anomaly data, 1993.

\bibitem{suthaharan2010labelled}
Shan Suthaharan, Mohammed Alzahrani, Sutharshan Rajasegarar, Christopher
  Leckie, and Marimuthu Palaniswami.
\newblock Labelled data collection for anomaly detection in wireless sensor
  networks.
\newblock In {\em 2010 sixth international conference on intelligent sensors,
  sensor networks and information processing}, pages 269--274. IEEE, 2010.

\bibitem{mafaulda}
Felipe M.~L. Ribeiro.
\newblock Machinery fault database.
\newblock {\em Signals, Multimedia, and Telecommunications Laborator, UFRJ.},
  2016.

\bibitem{katser2020skoltech}
Iurii~D Katser and Vyacheslav~O Kozitsin.
\newblock Skoltech anomaly benchmark (skab).
\newblock {\em Kaggle}, 2020.

\bibitem{ref_article3}
Iurii~D Katser, Vyacheslav~O Kozitsin, Ivan~V Maksimov, Denis~A Larionov, and
  Konstantin~I Kotsoev.
\newblock Data pre-processing methods for npp equipment diagnostics algorithms:
  an overview.
\newblock {\em Nuclear Energy and Technology}, 7:111, 2021.

\bibitem{christ2018time}
Maximilian Christ, Nils Braun, Julius Neuffer, and Andreas~W Kempa-Liehr.
\newblock Time series feature extraction on basis of scalable hypothesis tests
  (tsfresh--a python package).
\newblock {\em Neurocomputing}, 307:72--77, 2018.

\bibitem{ref_article4}
Guolin Ke, Qi~Meng, Thomas Finley, Taifeng Wang, Wei Chen, Weidong Ma, Qiwei
  Ye, and Tie-Yan Liu.
\newblock Lightgbm: A highly efficient gradient boosting decision tree.
\newblock {\em Advances in neural information processing systems}, 30, 2017.

\bibitem{ref_article5}
Fabian Pedregosa, Ga{\"e}l Varoquaux, Alexandre Gramfort, Vincent Michel,
  Bertrand Thirion, Olivier Grisel, Mathieu Blondel, Peter Prettenhofer, Ron
  Weiss, Vincent Dubourg, et~al.
\newblock Scikit-learn: Machine learning in python.
\newblock {\em the Journal of machine Learning research}, 12:2825--2830, 2011.

\bibitem{ref_article6}
Mart{\'\i}n Abadi, Ashish Agarwal, Paul Barham, Eugene Brevdo, Zhifeng Chen,
  Craig Citro, Greg~S Corrado, Andy Davis, Jeffrey Dean, Matthieu Devin, et~al.
\newblock Tensorflow: Large-scale machine learning on heterogeneous distributed
  systems.
\newblock {\em arXiv preprint arXiv:1603.04467}, 2016.

\bibitem{ref_article7}
Tianqi Chen and Carlos Guestrin.
\newblock Xgboost: A scalable tree boosting system.
\newblock In {\em Proceedings of the 22nd acm sigkdd international conference
  on knowledge discovery and data mining}, pages 785--794, 2016.

\bibitem{grandini2020metrics}
Margherita Grandini, Enrico Bagli, and Giorgio Visani.
\newblock Metrics for multi-class classification: an overview.
\newblock {\em arXiv preprint arXiv:2008.05756}, 2020.

\end{thebibliography}

\end{document}